\documentclass[11pt,logo,onecolumn,copyright]{nvidiatechreport}
\usepackage[round]{natbib}
\usepackage{xspace}
\usepackage{placeins}
\usepackage{float}
\usepackage[most]{tcolorbox}
\usepackage{array}
\usepackage{ragged2e}
\usepackage{microtype}
\usepackage{multirow}
\definecolor{NvidiaGreen}{HTML}{76B900}
\definecolor{DraftGray}{HTML}{56616A}
\newcommand{\ourmodel}{Nemotron 3.5 Bolt 74B-A9B 24GB\xspace}
\newcommand{\parentmodel}{Nemotron 3.5 Super \xspace}

\definecolor{TodoRed}{HTML}{B42318}
\newcommand{\nametodo}[1]
{\par\noindent{\small\color{TodoRed}\textbf{#1\if\relax\detokenize{#1}\relax\else: \fi TODO}}\par}
\newcommand{\draftfigure}[2]{\IfFileExists{figures/#1.pdf}{\includegraphics[width=\linewidth]{figures/#1.pdf}}{\fbox{\parbox[c][4cm][c]{0.95\linewidth}{\centering Figure in preparation: #2}}}}
\setlist[itemize]{topsep=2pt,itemsep=2pt}
\renewcommand{\today}{15 October 2026}
\title{Nemotron 3.5 Bolt 74B-A9B 24GB:\\A Hybrid Vision-Language Model for Local AI}
\author{\large NVIDIA}
\date{}
\begin{document}
\newcommand{\tb}[1]{\textcolor{blue}{[TB: #1]}}
\newcommand{\spr}[1]{\textcolor{red}{[SP: #1]}}
\newcommand{\alit}[1]{\textcolor{magenta}{[AT: #1]}}
\newcommand{\jl}[1]{\textcolor{cyan}{[JL: #1]}}
\newcommand{\oo}[1]{\textcolor{teal}{[OO: #1]}}
\begin{abstract}
\textbf{Abstract.} We introduce \ourmodel, a hybrid Mamba--Transformer vision-language model that optimizes speed and intelligence per GB of memory used for local agentic workloads. Derived from the \parentmodel VL 120B-A12B model family, Nemotron 3.5 Bolt 74B-A9B 24GB combines Star Elastic compression with 2-bit Quantization-Aware Distillation for memory-constrained deployment. The model is optimized for reasoning, agentic tool use, and web \& terminal interactions. We also trained a LiLiCorr speculative drafter that is released alongside the model to reach our target inference speed. The resulting model achieves a higher accuracy than leading models with similar memory footprints and speed across many benchmarks for general agentic capabilities (e.g., GDPVal v2), web browsing (e.g., BrowseComp), and terminal use (e.g., Terminal-Bench 2.1), with a relatively high TPS of 84 on RTX Spark that is comparable to models with much lower active parameter counts and more than sufficient for long-horizon agentic applications.
\end{abstract}
\maketitle
\begin{center}
\end{center}
\section{Introduction}
\label{sec:introduction}
A compelling local agent experience requires a memory-efficient language model that can run alongside other user applications, and optimizes speed and intelligence per GB of memory used.

We present \ourmodel, a hybrid mixture-of-experts vision-language model designed for reasoning, multimodal understanding, and local agentic workloads on NVIDIA RTX Spark \& DGX Spark. It fits into 24GB of memory and achieves a competitive decoding speed of 84 TPS on RTX Spark, while maximizing intelligence on benchmarks relevant to local agents: general agentic capabilities, web browsing, and terminal use. This is achieved 
through a combination of model compression, post-training, multimodal adaptation, low-bit quantization, and speculative decoding.

We begin from a \parentmodel VL 120B A12B model checkpoint and use Star Elastic~\citep{taghibakhshi2026starelastic_arxiv} to construct a compact backbone for our deployment budget. Guided by the deployment budget plus Knowledge Distillation (KD) signals, a router selects the routed-expert count and MoE intermediate width for each MoE layer, while preserving the hybrid architecture and hidden size to retain compatibility with the parent’s vision head. KD from the parent jointly trains the router and student weights, while a budget loss guides the architecture choices. Once the router converges on a candidate architecture that meets the target budget, we continue KD at short and long contexts to recover the parent's accuracy.

Further post-training then addresses instruction following, model behavior, and agentic task performance through SFT and reinforcement learning. These stages improve the compact model on the capabilities needed for local assistants and long-running agents. The recovery work addresses tool-use and software-engineering tasks, together with multimodal alignment of the vision interface to the compressed language backbone. Evaluation covers text reasoning and coding as well as visual grounding, screen understanding, and browser interaction. 

We further compress the network with low-bit Quantization-Aware Distillation ~\citep{xin2026quantizationawaredistillationnvfp4inference}, which quantizes MoE routed and shared experts to an aggressive 2.0625 bits, leveraging NVFP4 for other layers. This brings the model size from a 148 GB BF16 compressed model to a 22.1 GB quantized model, an 85.1\% reduction in size. This quantization is lossy by design. Our ablations show it is more Pareto-optimal to quantize a larger network to lower bits, than to have a smaller model at a higher bit-width, leading to a much stronger model for its size.

Lastly, we use LiLiCorr ~\citep{rusanovsky2026lilicorrlightweightlikelihoodcorrelation} for our speculative-decoding approach, improving interactive generation. Runtime integration covers llama.cpp and vLLM, with deployment accounting for model weights, the vision stack, the drafter, context state, and execution workspaces. Altogether, we made the full system practical under the memory and latency constraints of Local AI inference.

\begin{center}
\includegraphics[width=0.8\linewidth]{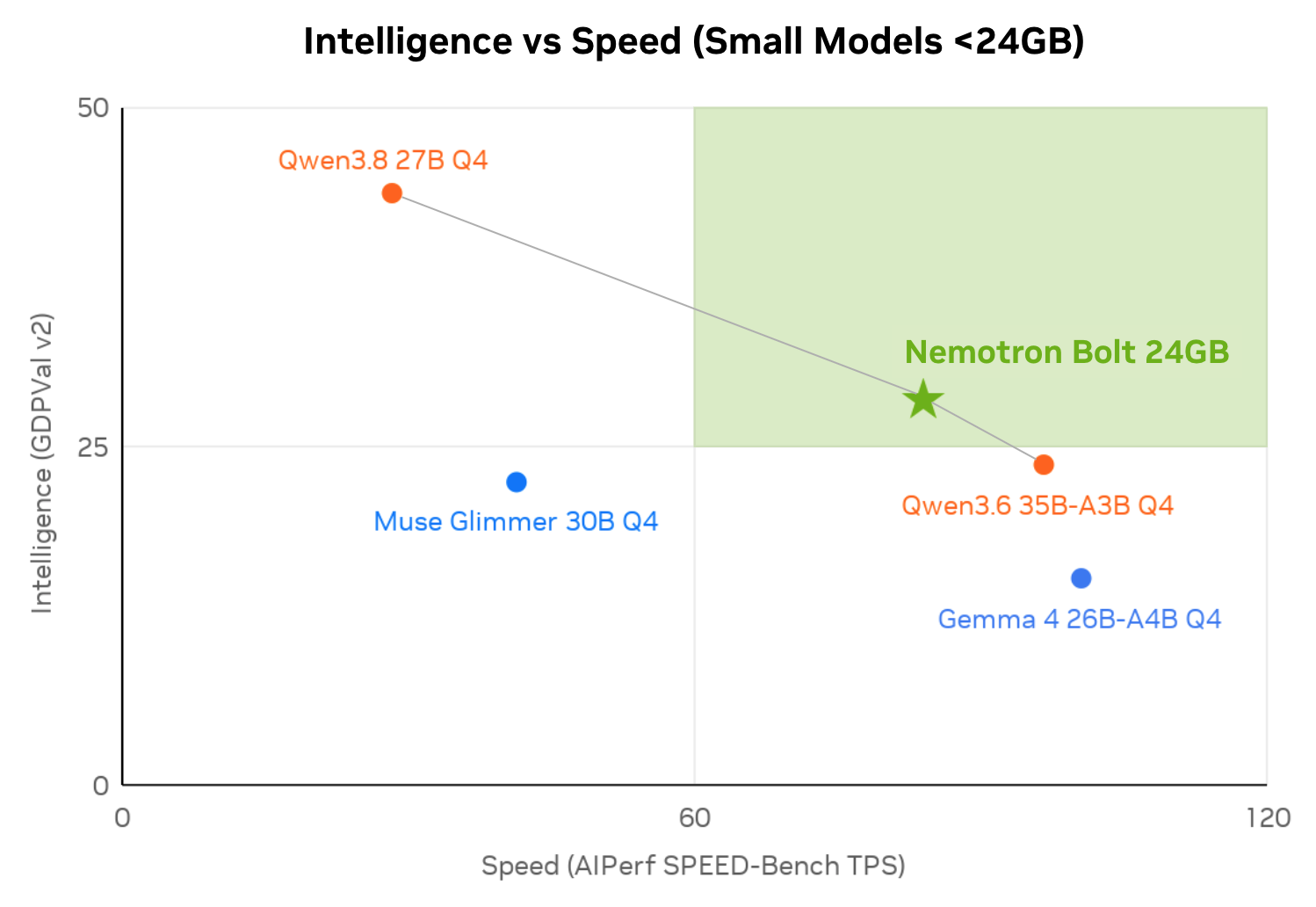}
\end{center}

\paragraph{Checkpoints}
\begin{itemize}
    \item \href{https://huggingface.co/nvidia/NVIDIA-Nemotron-3.5-Bolt-74B-A9B-24GB}{\texttt{NVIDIA-Nemotron-3.5-Bolt-74B-A9B-24GB} \includegraphics[height=0.9em]{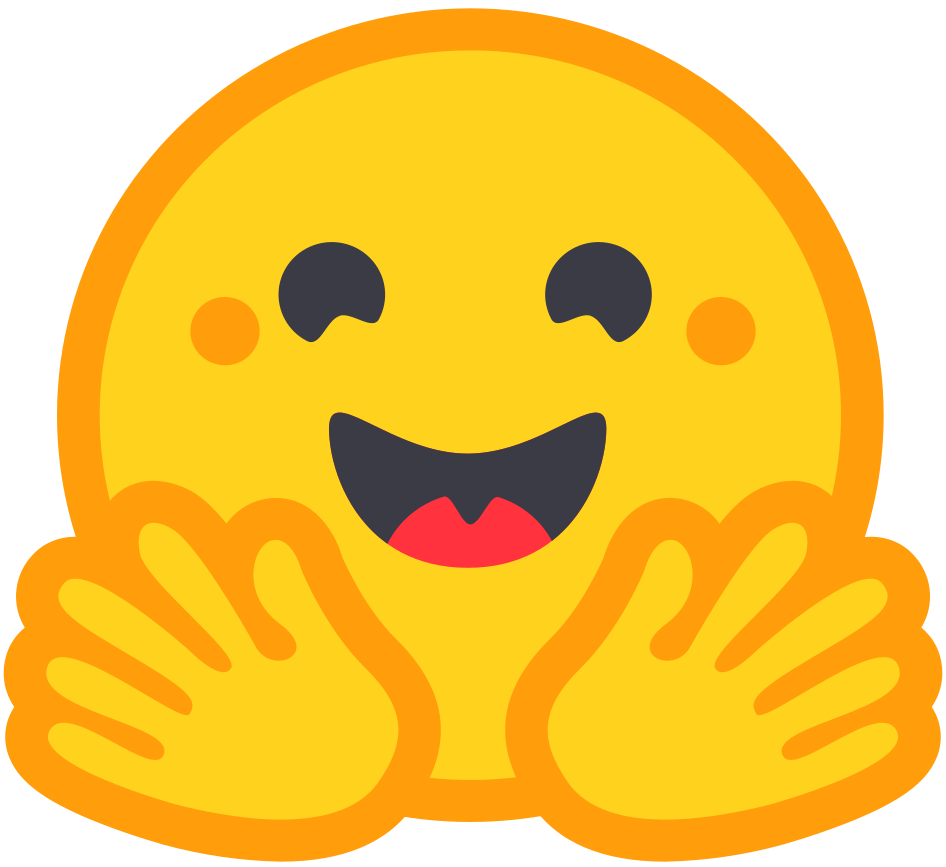}}:
    IQ2-quantized model in Hugging Face format.

    \item \href{https://huggingface.co/nvidia/NVIDIA-Nemotron-3.5-Bolt-74B-A9B-BF16}{\texttt{NVIDIA-Nemotron-3.5-Bolt-74B-A9B-BF16} \includegraphics[height=0.9em]{figures/huggingface-color.png}}:
    BF16 model in Hugging Face format.

    \item \href{https://huggingface.co/nvidia/NVIDIA-Nemotron-3.5-Bolt-74B-A9B-24GB-GGUF}{\texttt{NVIDIA-Nemotron-3.5-Bolt-74B-A9B-24GB-GGUF} \includegraphics[height=0.9em]{figures/huggingface-color.png}}:
    IQ2-quantized model in GGUF format.

    \item \href{https://huggingface.co/nvidia/NVIDIA-Nemotron-3.5-Bolt-74B-A9B-24GB-LiLiCorr}{\texttt{NVIDIA-Nemotron-3.5-Bolt-74B-A9B-24GB-LiLiCorr} \includegraphics[height=0.9em]{figures/huggingface-color.png}}:
    LiLiCorr drafter for the IQ2-quantized model, in Hugging Face format.

    \item \href{https://huggingface.co/nvidia/NVIDIA-Nemotron-3.5-Bolt-74B-A9B-24GB-LiLiCorr-GGUF}{\texttt{NVIDIA-Nemotron-3.5-Bolt-74B-A9B-24GB-LiLiCorr-GGUF} \includegraphics[height=0.9em]{figures/huggingface-color.png}}:
    LiLiCorr drafter for the IQ2-quantized model, in GGUF format.
\end{itemize}
\section{Compression}
\label{sec:compression}

We compress the Nemotron 3.5 Super VL parent checkpoint (120B-A12B) to Nemotron 3.5 Bolt (74B-A9B) using Star Elastic~\citep{taghibakhshi2026starelastic_arxiv}, NVIDIA's framework for joint architecture search and model compression.

\subsection{Architecture search}
We derive a target parameter count $b$ from the deployment memory budget, accounting for the intended quantization scheme and runtime-state precisions. A budget-conditioned architecture router takes an encoding of $b$ and selects the routed-expert count and MoE intermediate width for each MoE layer. The router has a separate MLP head for each axis. Each head consists of two linear layers with a LeakyReLU activation between them and outputs a score for every candidate size in every MoE layer. Expert top-$K$ remains fixed throughout the search.

Before the search, using a calibration dataset obtained from a subset of the training data, we sort the routed experts and MoE intermediate channels in each MoE layer in descending order using the scoring mechanisms described in REAP~\citep{lasby2026reap} and Star Elastic~\citep{taghibakhshi2026starelastic_arxiv}, respectively.  Reordering experts also reorders the corresponding outputs of the token router; reordering channels permutes the up-projection rows and matching down-projection columns. These permutations preserve the parent model's function before masking.

\paragraph{Differentiable masks.}

For every MoE layer and axis, we define $c_j$ to be the $j$th candidate size among $C$ choices. In the sorted parent checkpoint, the binary prefix mask for candidate $j$ is
\begin{equation}
m_j[i]=\mathbf{1}\!\left\{i\leq c_j\right\},
\label{eq:prefix-mask}
\end{equation}
where $i$ indexes sorted experts or channels, starting at 1, and $\mathbf{1}\{\cdot\}$ is the indicator function. Smaller masks therefore retain a subset of the components kept by larger masks.

\begin{figure}[!htbp]
\centering
\captionsetup{skip=4pt}
\includegraphics[width=0.94\linewidth]{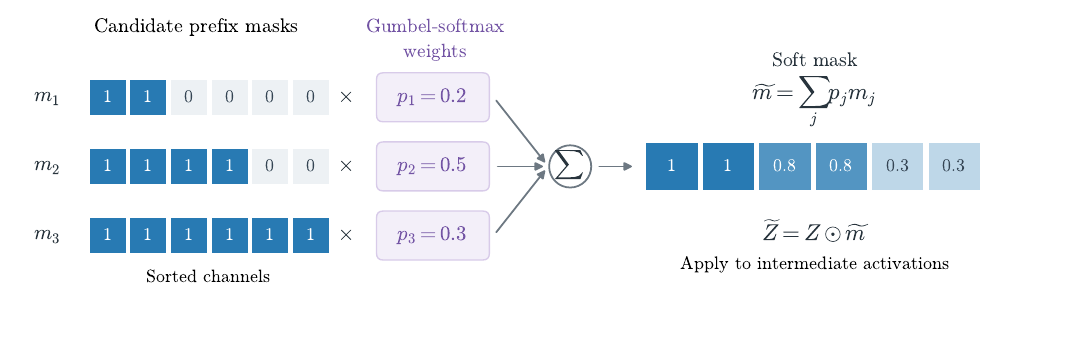}
\caption{Illustration of soft masking for one layer and axis. Gumbel-softmax weights combine nested prefix masks; the dashed arrow shows the gradient path.}
\label{fig:soft-masking}
\end{figure}

Following Star Elastic~\citep{taghibakhshi2026starelastic_arxiv}, we use the Gumbel-softmax~\citep{jang2016categorical} trick on top of the architecture router's scaled logits. Multiplying the Gumbel logits by the prefix mask makes the masking process differentiable, and that mask is multiplied by the corresponding layer's activation, letting the distillation loss change the mixture of candidate masks through backpropagation. We use a soft masking procedure, where all the mask choices are applied to the corresponding activation, and a weighted average is taken according to the Gumbel logits, resulting in a soft forward pass until the router converges. Figure~\ref{fig:soft-masking} illustrates the mixture for one layer and axis.

\paragraph{Knowledge-distillation loss.}
We use the frozen parent checkpoint as teacher $T$ and minimize the token-level KL divergence between its next-token distribution and the student's.

\paragraph{Router budget loss.}
The deployment constraint is memory. For architecture search, we
translate this constraint into a target parameter count $b$, using
the intended quantization recipe, runtime-state precisions, and desired workload (sequence length, concurrency, etc.). For more details on how this translation is done, please refer to Section~\ref{sec:checkpoint-size}.
Using this conversion,
we express the budget loss in parameter space, penalizing deviations
from the corresponding target $b$ beyond a tolerance.
Let $P(\psi;b)$ denote the differentiable
parameter-count estimate obtained from the router-weighted
layer capacities, where $\psi$ denotes the router's trainable parameters. The relative error $\delta$ and budget loss are
\begin{equation}
\delta=
\left|\frac{ P(\psi;b)}{b}-1\right|,
\qquad
\mathcal L_{\mathrm{budget}}(\psi;b)=
\begin{cases}
0, & \delta\leq\epsilon,\\
\delta, & \delta>\epsilon,
\end{cases}
\label{eq:compression-budget}
\end{equation}
where $\epsilon$ is the relative budget tolerance. This loss updates only the architecture router parameters $\psi$.

The joint objective is
\begin{equation}
\min_{\theta,\psi}\;
\mathcal L_{\mathrm{KD}}(\theta,\psi;b)
+\lambda\,\mathcal L_{\mathrm{budget}}(\psi;b),
\label{eq:objective}
\end{equation}
where $\theta$ denotes the student model's parameters, and $\lambda\geq0$ sets the weight of the budget loss relative to distillation. Both losses train the architecture router; only distillation trains the student weights. 

\begin{figure}[!htbp]
\centering
\captionsetup{skip=4pt}
\includegraphics[width=\linewidth]{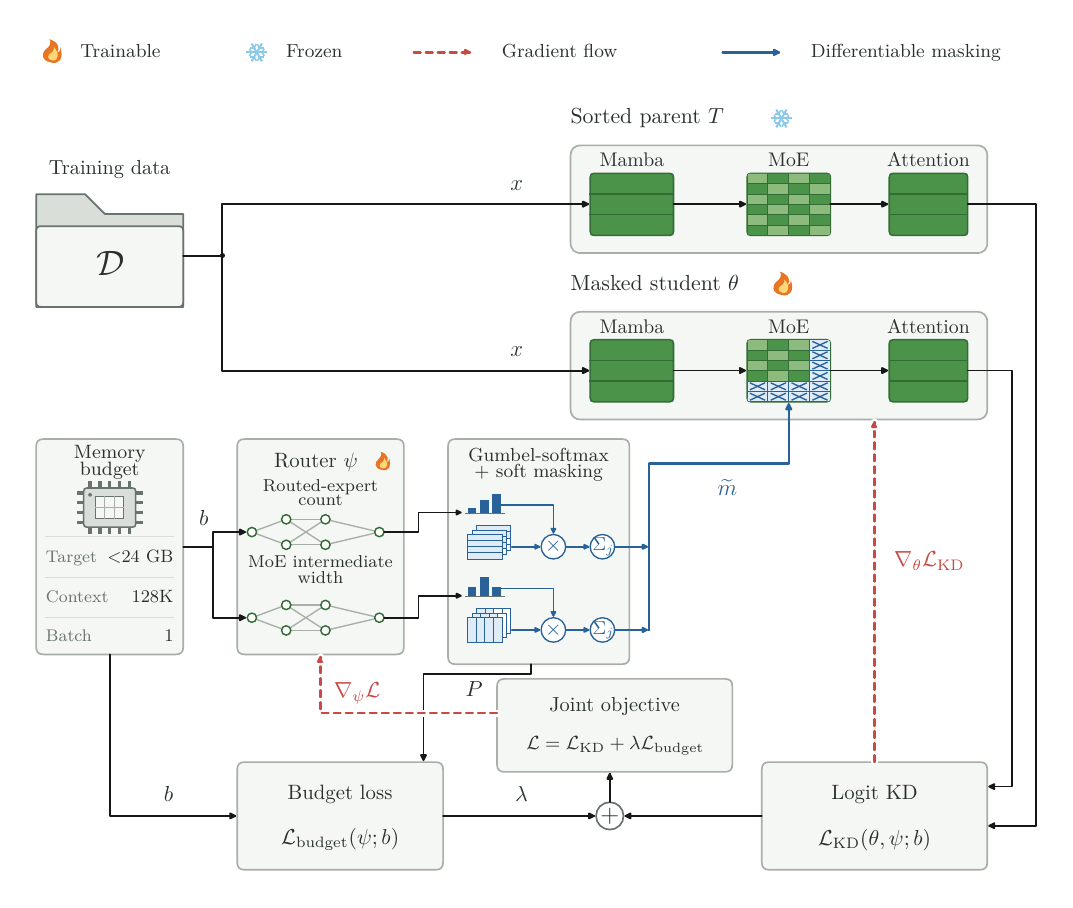}
\caption{Star Elastic compression pipeline. Architecture search and knowledge distillation jointly optimize the student weights, routed-expert counts, and MoE intermediate widths. A budget penalty guides the architecture router toward the deployment memory target. See~\citep{taghibakhshi2026starelastic_arxiv} for full details.}
\label{fig:search}
\end{figure}

\subsection{Obtained architecture}
\label{sec:architecture}
\begin{figure}[!htbp]
\centering
\captionsetup{skip=4pt}
\includegraphics[width=\linewidth]{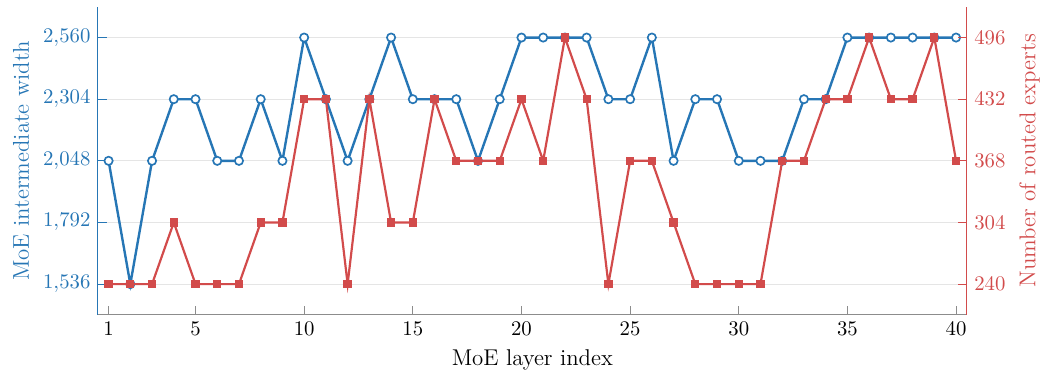}
\caption{MoE intermediate width and number of routed experts across MoE layers, ordered by backbone depth.}
\label{fig:architecture}
\end{figure}

The compressed model retains the parent's Mamba, MoE, and attention layers and all dimensions outside the routed experts (Table~\ref{tab:architecture}). Routed-expert counts and MoE intermediate widths vary across layers, as shown in Figure~\ref{fig:architecture}. Experts within a layer share the same intermediate width.

\begin{table}[!htbp]
\centering
\small
\renewcommand{\arraystretch}{1.15}
\begin{tabularx}{\linewidth}{@{}>{\RaggedRight\arraybackslash}p{0.38\linewidth} >{\centering\arraybackslash}X >{\centering\arraybackslash}X@{}}
\toprule
\textbf{Property} & \textbf{Nemotron 3.5 Super VL parent} & \textbf{Nemotron 3.5 Bolt} \\
\midrule
Mamba / MoE / attention layers & 40 / 40 / 8 & 40 / 40 / 8 \\
Hidden width & 4096 & 4096 \\
Latent MoE width & 1024 & 1024 \\
Shared-expert intermediate width & 5376 & 5376 \\
Routed experts per MoE layer & 512 & $\in\{240,304,368,432,496\}^{*}$ \\
MoE intermediate width & 2688 & $\in\{1536,2048,2304,2560\}^{*}$ \\
Experts selected per token & 22 & 9 \\
\bottomrule
\end{tabularx}
\caption{Parent and obtained architecture. $^{*}$Each MoE layer's routed-expert count and intermediate width belong to their respective sets.}
\label{tab:architecture}
\end{table}

\subsection{Compression settings}
\label{sec:compression-setting}
The deployment target is below 24\,GB at batch size 1 and 128K context. The memory estimate includes the intended weight and runtime-state precisions, including 2-bit routed-expert weights and 16-bit SSM states. Section~\ref{sec:quantization} and Table~\ref{tab:bolt-tensor-precision} give the full precision configuration. We fix expert top-$K$ at 9, down from 22 in the parent.

We train the architecture router for a single budget. Each head has dimensions $1\rightarrow r_D\rightarrow L C_a$, where $r_D=256$ is the hidden width, $L=40$ is the number of MoE layers, and $C_a$ is the number of candidate sizes for axis $a$ (expert count or intermediate width). The LeakyReLU negative slope is 0.1. We use a router learning rate of $5\times10^{-3}$, budget-loss weight $\lambda=1$, and relative budget tolerance $\epsilon=0.05$.

For calibration, search, and recovery, we use a blend of 70\% SFT data used to train the Nemotron 3.5 Super VL parent and 30\% Nemotron Ultra-v3 pretraining data. We estimate expert and channel importance on 8M tokens sampled from this shuffled blend.

Architecture search runs at 8K context, with the router converging during the first 8B tokens. We then fix the architecture and continue distillation for another 64B tokens at 8K, for a short-context total of 72B tokens. A further 48B tokens at 128K brings the total to 120B tokens. Following Star Elastic's schedule, we perform most recovery at 8K, where it proceeds faster, before extending to long contexts.

Training used 32 GB200 nodes (128 GPUs) for approximately 64 hours, across 16 consecutive four-hour jobs. The global batch size was 2,048 sequences at 8K and 128 sequences at 128K, giving approximately 16.8M tokens per batch in both settings.

\subsection{Compression quality}
\label{sec:compression-quality}
\begin{figure}[!htbp]
\centering
\captionsetup{skip=4pt}
% Trim blank PDF margins: left, bottom, right, top.
\includegraphics[width=\linewidth,trim={8bp 20bp 4bp 5bp},clip]{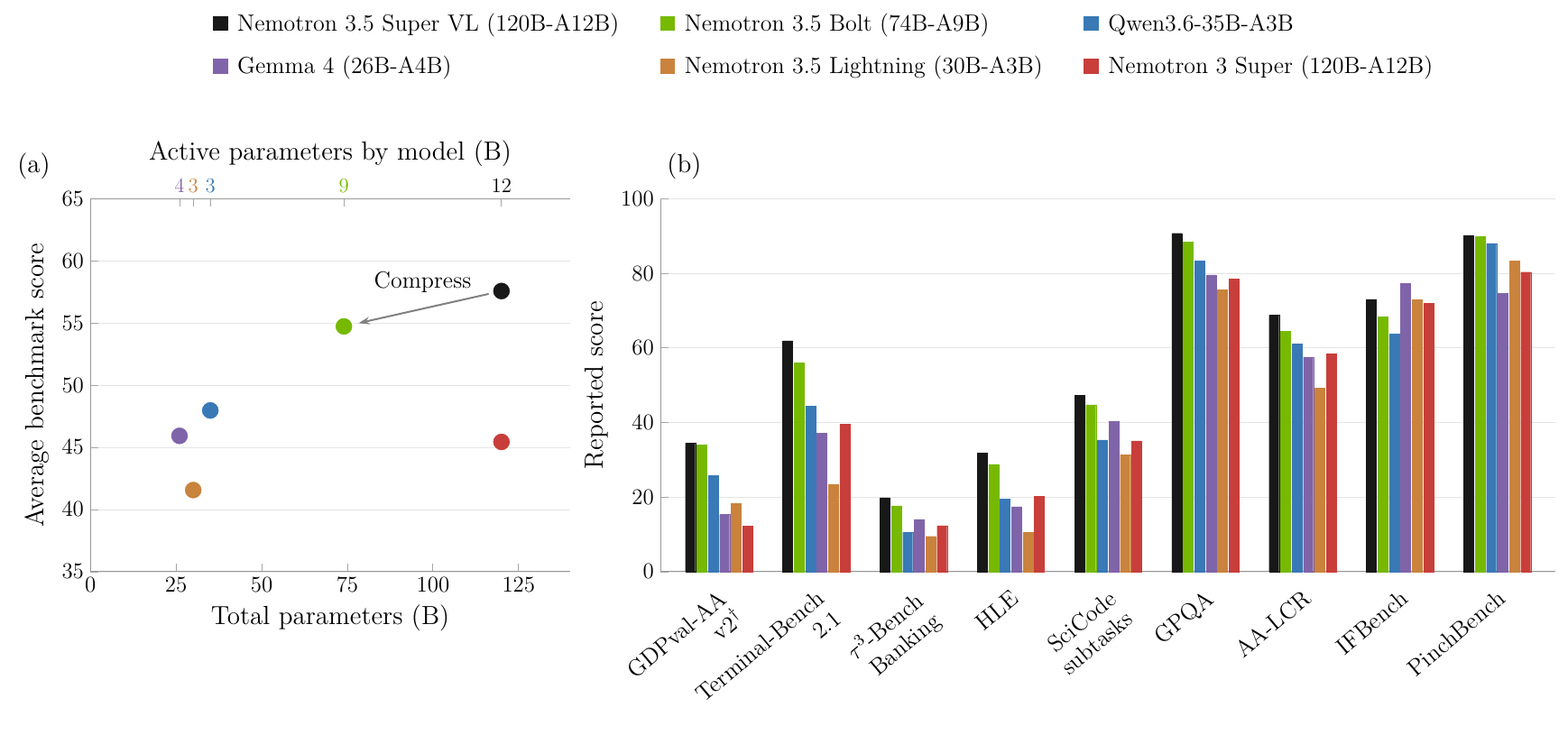}
\caption{Per-benchmark performance of Nemotron 3.5 Bolt compared with the Nemotron 3.5 Super VL parent and community models. (a) Mean score across nine benchmarks versus total parameters (bottom axis); upper ticks give each model's active parameter count at its total-parameter position. Both counts are in billions. The arrow marks compression from the parent to Nemotron 3.5 Bolt. (b) Individual benchmark scores. $\dagger$~GDPval-AA v2 is an index; the other scores are percentages.}
\label{fig:compression-parameter-comparison}
\label{fig:compression-memory-comparison}
\label{fig:compression-benchmarks}
\end{figure}

Figure~\ref{fig:compression-parameter-comparison} compares Nemotron 3.5 Bolt with the Nemotron 3.5 Super VL parent checkpoint (120B-A12B) and four reference models. Across the nine reported metrics, the unweighted mean decreases from 57.58 for the parent to 54.74 for Nemotron 3.5 Bolt, a loss of 2.85 points, or 4.94\% relative to the parent. The parent scores are from BF16 evaluation, and the Nemotron 3.5 Bolt scores precede additional SFT/RL healing and quantization. These scores do not measure the accuracy of the quantized deployment configurations.

\clearpage
\section{Multimodal Adaptation}
\label{sec:multimodal}

We constructed the Nemotron 3.5 Bolt 74B-A9B VL model by re-attaching the trained vision tower from the parent Super 3.5 VL model back to the compressed LLM. We then performed lightweight fine-tuning of only the vision-to-language projection layers (the vision adapter), using image-text datasets previously used to train the parent model’s vision adapter.

Training used an image-only blend of approximately 6.6M samples per epoch. The blend comprised a general English mix of captioning, visual question answering, counting, and document datasets (\textasciitilde2M samples), CC3M images recaptioned by Qwen3-30B (\textasciitilde1.7M samples), and Chinese document OCR data (\textasciitilde2.5M samples). It also included PDF and table document data, along with datasets used to train NVCLIP OCR. No video or text-only data was used during vision adapter training.

\section{SFT and RL Healing After Compression}

\begin{figure}[htbp]
\centering
\includegraphics[width=\linewidth]{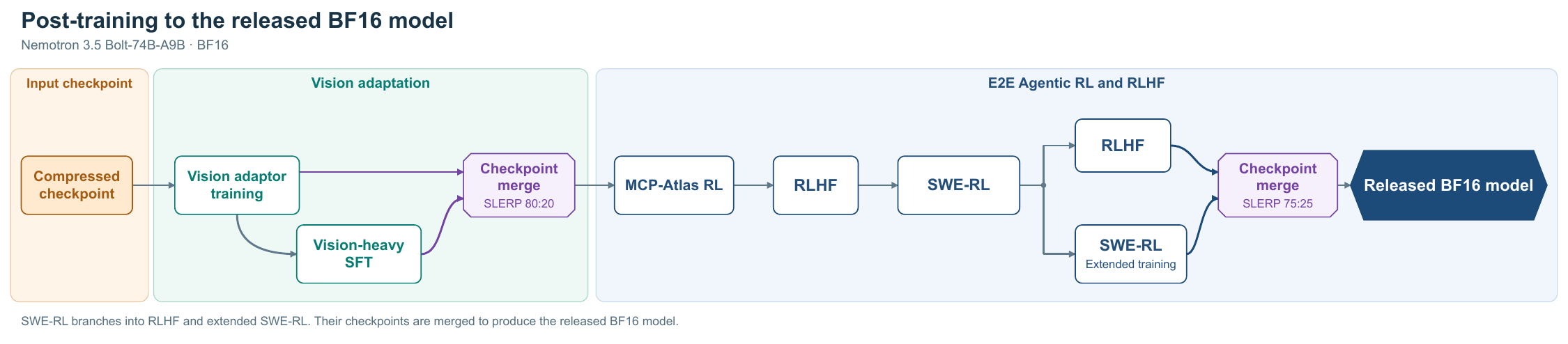}
\caption{Post-training journey after compression}
\label{fig:posttrain}
\end{figure}

Figure~\ref{fig:posttrain} details all training and merging stages from the checkpoint compressed with Star Elastic to the final BF16 checkpoint.

Post vision-adaptor training, the model is trained with vision-heavy SFT training to recover vision capabilities further. Before moving to multi-turn agentic training, we applied checkpoint merging as it provided more well-rounded capabilities across text and vision capabilities.

The model underwent further training in agentic environments and with RLHF, followed by a final checkpoint-merging step to produce the BF16 release checkpoint. We describe each training stage in greater technical detail below

\label{sec:healing}
\subsection{Supervised Recovery}
\label{subsec:SFT_stage}
We used the same SFT data blends as those used to train the parent Super 3.5 VL model, but reweighted them to emphasize vision. The original blend contained approximately 54.1B effective tokens across 7.3M samples and was slightly text-majority, with 47.5\% vision tokens and 52.5\% text tokens.

For vision, we used dataset blends targeting ScreenSpot Pro ($\sim$10.1B tokens) and WebArena ($\sim$9.5B tokens), supplemented with small amounts of MolmoWeb/WebVoyager browsing data, CharXiv chart data, HopChain multi-page document data, and BabyVision-targeted reasoning data. For text, we used a downsampled blend of Kimi-K3-generated synthetic datasets targeting the AIME, GPQA Diamond, Terminal Bench 2.1, SciCode, AA-LCR, APEX, and LMArena benchmarks. We also incorporated a very small proportion of GLM 5.2-generated trajectories targeting MCP-Atlas. During this stage, the LLM, vision-adaptor and the vision encoder were all trained.

Following the vision-heavy SFT stage, we performed a model merge with the vision adaptor only trained checkpoint using SLERP interpolation. 

\begin{figure}[htbp]
\centering
\includegraphics[width=\linewidth]{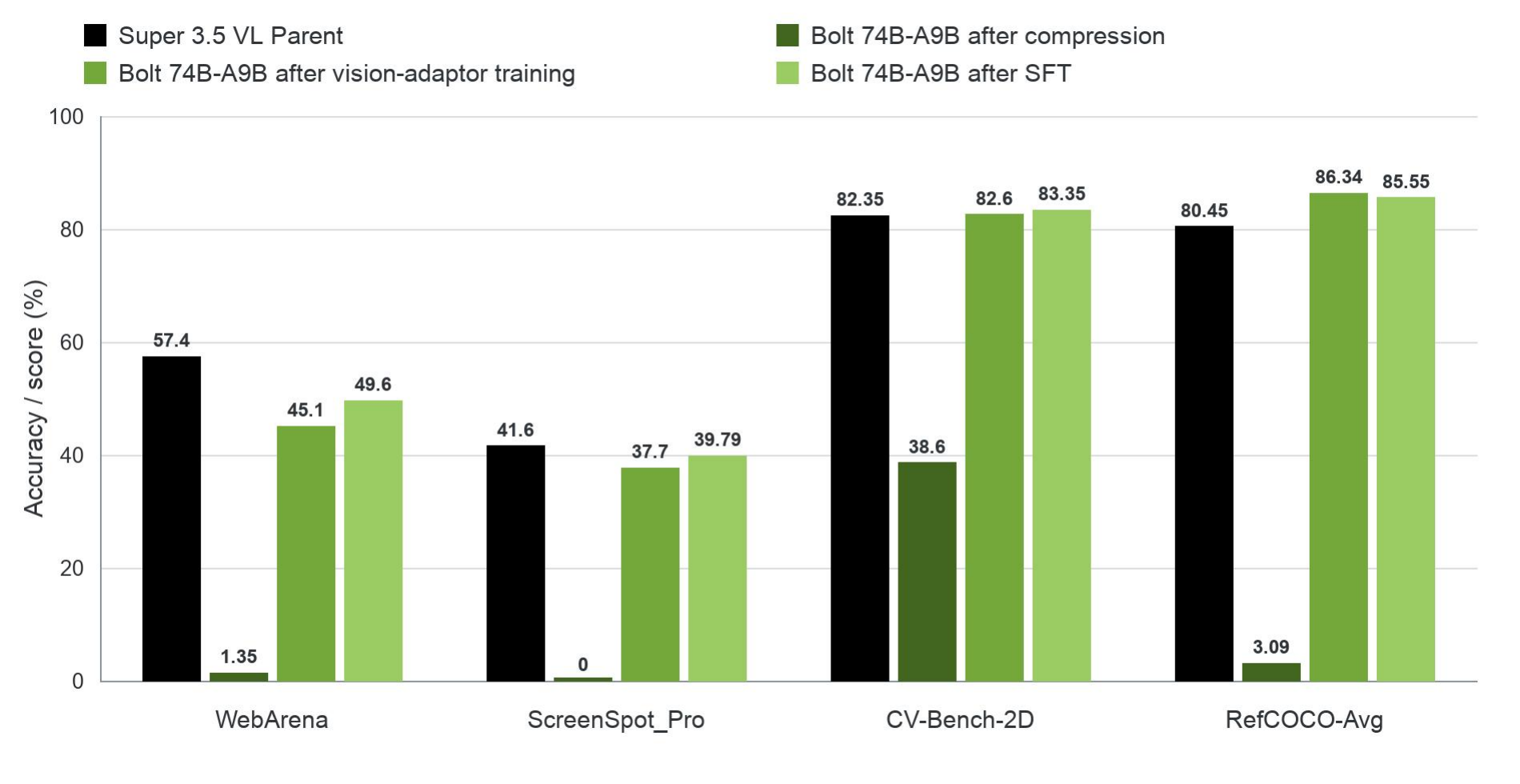}
\caption{Vision benchmarks recovery after compression}
\label{fig:vision_adaptor_sft}
\end{figure}

Vision adaptor training and an SFT stage with a vision heavy dataset (see Figure~\ref{fig:vision_adaptor_sft}) is critical to recovering vision benchmarks, as compression was performed only on the text tower of the parent model with knowledge distillation applied to recover text-only benchmarks during compression. There is little to no degradation in text accuracy.

\subsection{Reinforcement Learning}
\label{subsec:RL_stage}

The model was further trained with RLHF and several stages of E2E RL training to improve agentic capabilities and minimize task completion times for tool-heavy multi-turn workloads. 

During RLHF, we used the Nemotron 3 Ultra GenRM model which was also trained with a reasoning penalty for token efficiency.

For further training, we have used multiple agentic environments with full E2E GRPO to target MCP and SWE tasks.  Furthermore, we used Terminal Environment to generate on-policy rollouts to prevent regressions during quantization, although it was not used in BF16 training. 
    
Our MCP E2E RL environment includes a simulated MCP-server with tasks requiring multi-turn tool interactions to achieve the goals. We used 32 prompts with 16 generations per prompt, giving a batch size of 512. Training on this environment improved MCP use capabilities and provided an increase of around 8 nominal points from ~48 to ~56 on the 500 public tasks out of 1000 total tasks on the MCP-Atlas benchmark. The scoring mechanism for the MCP-Atlas benchmark uses a 0.75 threshold on coverage metric, i.e., the mean claim score (each task has multiple claims that are scored separately) and it provides a binary pass / fail score per task coverage in the end. We observed further RL training continued to provide improvement on coverage scores across tasks, but it did not translate to post-threshold based official score improvements. Whether longer RL training would have eventually provided improvements on the benchmark remains an open question. 

SWE-RL E2E training also used some optimizations to improve efficiency in terms of generated tokens and tool-call efficiency. Training used 64 prompts with 16 generations per prompt, giving 1,024 trajectories per optimizer step, with a maximum total sequence length of 196,608 tokens. We used Adam at a learning rate of $3 \times 10^{-6}$. This training also led to accuracy and efficiency improvements on PinchBench as a generalization benefit with scores up-to 90.4, averaged across 10-repeats.

\section{Quantization}

We quantize the experts of the model to 2-bits to fit the network in RTX Spark's memory. We use Quantization-Aware Distillation (QAD) ~\citep{xin2026quantizationawaredistillationnvfp4inference} to recover accuracy.
Quantization was performed using \mbox{ModelOpt}\footnote{\url{https://github.com/NVIDIA/Model-Optimizer}}.

\label{sec:quantization}
\subsection{Quantization Setup}

To further reduce the size of the network with quantization to 24GB, we quantize all experts to IQ2\_XXS, such that each parameter has 2.0625 bits per element ~\citep{llamacpp_tensor_encoding}. This includes both the routed and shared experts. Previous work on low-bit quantization such as ParetoQ ~\citep{liu2025paretoqimprovingscalinglaws} and Quest ~\citep{panferov2025queststabletrainingllms} has shown that lower-bits for FC layers could be pareto-optimal, i.e., layers at twice the size in 2 bits are better than the original size in 4 bits. This is a major lever we use to create our for-size checkpoint. In low-batch size inference scenarios, the latency of the kernels also improves linearly with the compression rate of the weights, leading to significant inference speed gains due to the low-bit quantization.

We performed further mixed-precision to achieve our memory constraint without impacting the ability of the model. The model keeps some layers quantized to either NVFP4 (Mamba layers) or Q8\_0 (Embedding and output layers). For a full overview, see Table \ref{tab:bolt-tensor-precision}: 

\begin{table}[!htbp]
\centering
\small
\renewcommand{\arraystretch}{1.12}
\begin{tabularx}{\linewidth}{@{}Xlrr@{}}
\toprule
\textbf{Component} &
\textbf{Stored format} &
\textbf{Bits} &
\textbf{Size (GB)} \\
\midrule
Attention projections          & \texttt{Q8\_0}    & 8.5    & 0.30 \\
Routed-expert projections       & \texttt{IQ2\_XXS} & 2.0625 & 17.05 \\
Shared-expert projections       & \texttt{IQ2\_XXS} & 2.0625 & 0.45 \\
Latent MoE projections          & \texttt{Q8\_0}    & 8.5    & 0.36 \\
Mamba input/output projections  & \texttt{NVFP4}    & 4.5    & 2.47 \\
Token embedding                & \texttt{Q8\_0}    & 8.5    & 0.57 \\
LM head                        & \texttt{Q8\_0}    & 8.5    & 0.57 \\
MoE router weights             & \texttt{F32}      & 32     & 0.23 \\
Mamba convolution weights      & \texttt{F32}      & 32     & 0.01 \\
\textbf{Weight subtotal}        &                   &        & \textbf{22.01} \\
\midrule
Mamba SSM cache                & \texttt{BF16}     & 16     & 0.08 \\
KV cache                       & \texttt{FP8}      & 8      & 0.54 \\
Activation buffer              & \texttt{Q8\_0}               & 8.5    & 0.25 \\
\textbf{Runtime subtotal}       &                   &        & \textbf{0.87} \\
\midrule
\textbf{Combined subtotal}      &                   &        & \textbf{22.87} \\
Speculative decoder             &  \texttt{NVFP4}/\texttt{BF16}                  &         & 0.96 \\
\textbf{Total excluding vision} &                   &        & \textbf{23.83} \\
\midrule
Vision encoder and projector   &   \texttt{BF16}                &        & \textbf{1.68} \\
\bottomrule
\end{tabularx}
\caption{Precision and estimated memory for Nemotron 3.5 Bolt.
Deployment settings are given in Section~\ref{sec:checkpoint-size}.}
\label{tab:bolt-tensor-precision}
\end{table}

Note that the quantization is very aggressive and lossy by design. The intent for this model is not to stay close to the model's BF16 accuracy. Rather, it is accuracy-wise more optimal to take as big a model as you can, and quantize it to lower-bits to get the best accuracy-for-size possible.

\subsection{Quantization-Aware Distillation}

Quantization-Aware Distillation has been a very effective method to recover accuracy for 4-bit deployments as described in ~\citep{xin2026quantizationawaredistillationnvfp4inference}. We extended QAD for 2-bit quantization. We further perform ablations to improve the accuracy gains as described below.

\paragraph {Training Schedule Ablation}
QAD requires careful learning rate selection, with different optimal ranges depending on the original training scheme. As described in \cite{xin2026quantizationawaredistillationnvfp4inference},
QAD benefits from a larger learning rate for agentic models. However, the optimal LR noted for NVFP4 formats has been around $5\times10^{-6}$, whereas the original SFT stage generally uses a higher learning rate. In line with the literature, we also observe that lower bit quantization requires an even higher learning rate, similar to or higher than even the SFT stage itself. From our ablations, we find that the best LR for 2-bit QAD was $1\times10^{-5}$ with warmup for 1.2B tokens and then cosine decay toward $1\times10^{-7}$ over 25B tokens.  

\paragraph {Data}
Previously, QAD for 4-bit models observed that the downstream evals were robust to the data blend being used, but in our aggressive quantization setup, we observed that the data blend plays a more important role in accuracy recovery.
We utilized the SFT data blend along with RL rollouts as described in section \ref{subsec:SFT_stage} and on-policy rollouts from the original BF16 model, used as a teacher. 

\paragraph{Loss Function Ablation}
Previous studies have explored multiple loss functions for QAD like KL-divergence ~\citep{hinton2015distillingknowledgeneuralnetwork}, Reverse KL-Divergence ~\citep{gu2026minillmonpolicydistillationlarge}, JSD loss ~\citep{wen-etal-2023-f}, MSE on logits, and cosine loss over intermediate layers ~\citep{sanh2020distilbertdistilledversionbert}. From our study, we have observed Jaccard loss to perform the best for low-bit QAD. We perform Quantization-Aware Distillation by minimizing a weighted Jaccard distance between the output distributions of the quantized student and a frozen teacher. At each training token, we compute the intersection and union of their vocabulary distributions using element-wise minima and maxima, respectively. The objective maximizes their normalized probability overlap and is averaged over tokens selected by the training loss mask. We use a softmax temperature of 1 and omit the supervised language-model cross-entropy term.

The Jaccard loss is
\begin{equation}
\mathcal{L}_{\mathrm{Jaccard}}
= 1 -
\frac{\sum_v \min\!\left(p_s(v), p_t(v)\right)}
     {\sum_v \max\!\left(p_s(v), p_t(v)\right)},
\end{equation}

where $p_s$ and $p_t$ are the student and teacher output
probabilities, and $v$ indexes the vocabulary.
We average the loss over valid training tokens.

\begin{figure}[htbp]
\centering
\includegraphics[width=\linewidth]{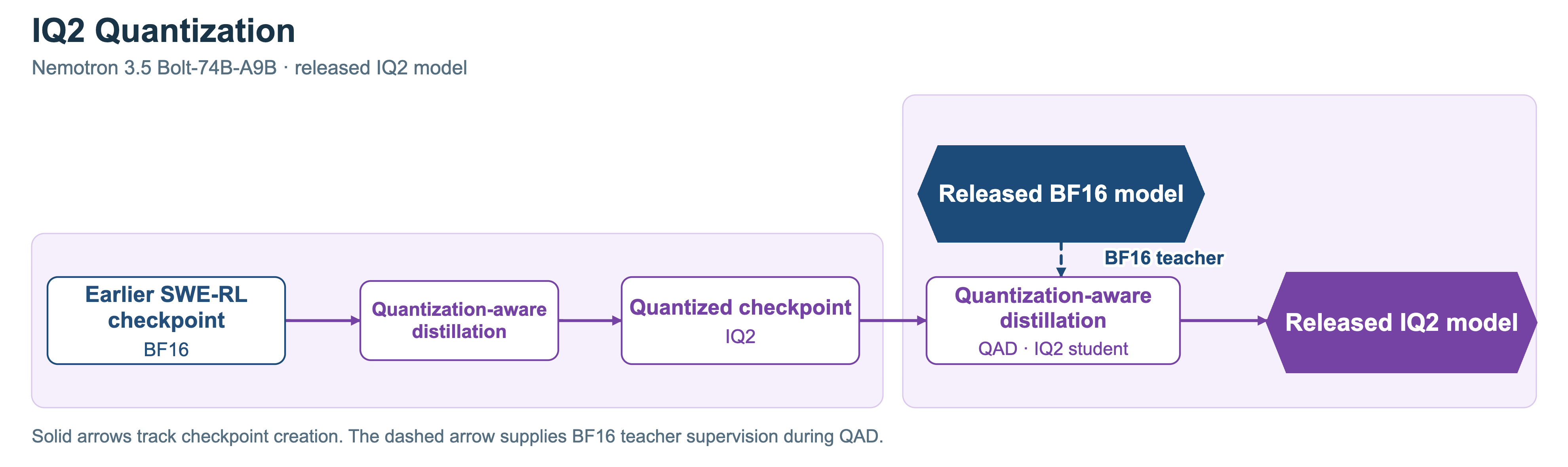}
\caption{Quantization-Aware-Distillation workflow}
\label{fig:qad}
\end{figure}

\paragraph{Using Multiple Teachers}

QAD was performed in 2 stages as shown in figure \ref{fig:qad}.

\begin{figure}[!htbp]
  \centering
  \includegraphics[width=0.9\linewidth]
  {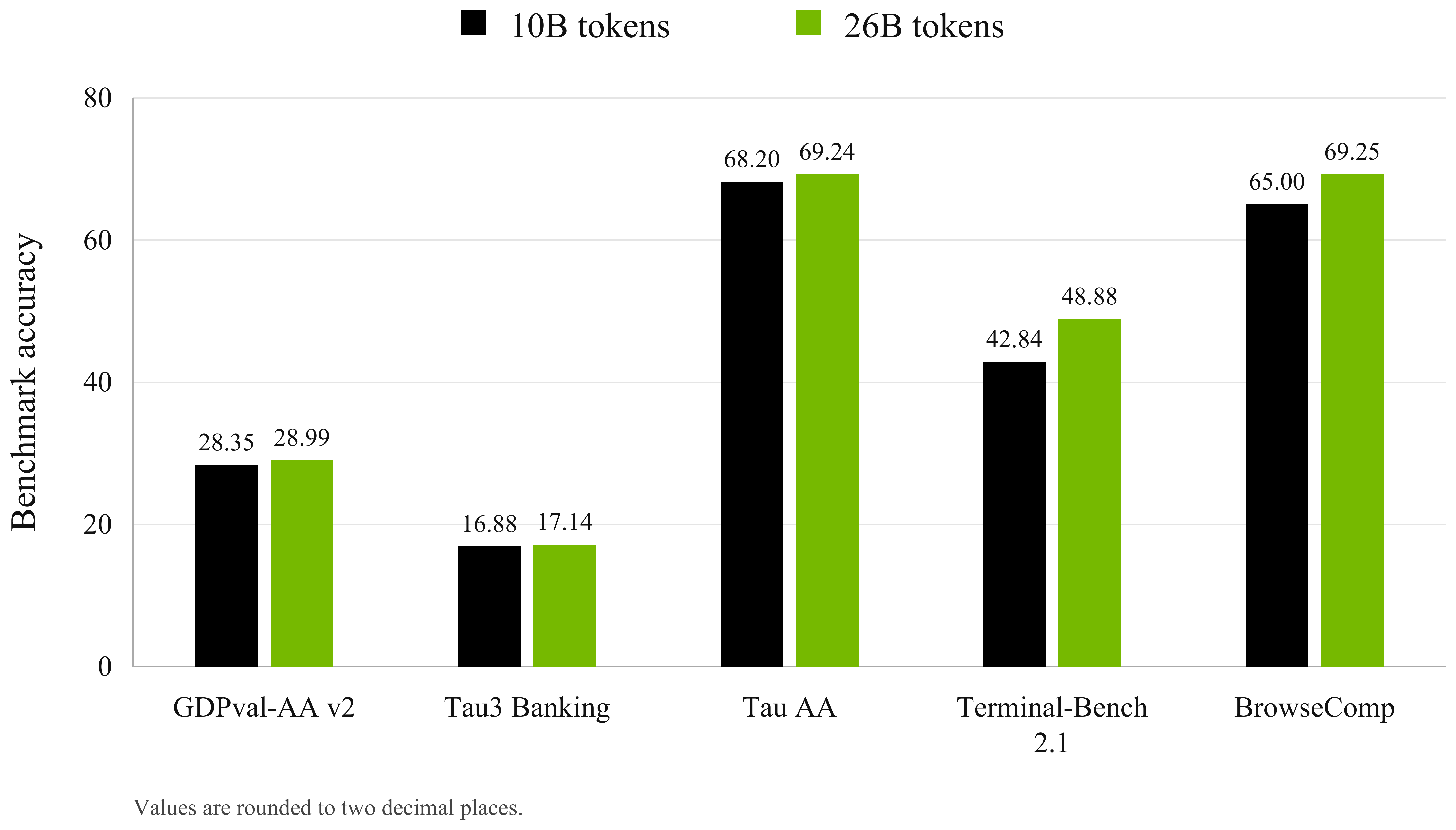}
  \caption{Benchmark accuracy after 10B and 26B training
  tokens. Longer quantization-aware training improves results across benchmarks.}
  \label{fig:longer_qad}
\end{figure}

The first stage of QAD used an earlier SWE-RL stage BF16 checkpoint, which had a very high accuracy. We performed QAD on 26B tokens to regain some of the accuracy lost due to quantization. Our ablations show that longer training helped regain accuracy across multiple benchmarks. The accuracy improvement over a longer token horizon can be seen in Figure \ref{fig:longer_qad}. 
After obtaining a strong IQ2\_XXS checkpoint, we utilized the final post-trained BF16 checkpoint that underwent additional RLHF and SWE-RL post-training, to improve the token efficiency of the quantized model. The second stage of QAD, using the newer, more efficient teacher, was run only for 10M tokens from the RLHF blend discussed in section \ref{subsec:RL_stage} to improve token and agentic efficiency. The final accuracy after quantization is reported in table \ref{tab:bolt-bf16-iq2-qwen-comparison}.

\subsection{Quantization quality}
\label{sec:quantization-quality}
\begin{figure}[!htbp]
  \centering
  \includegraphics[width=\linewidth]
  {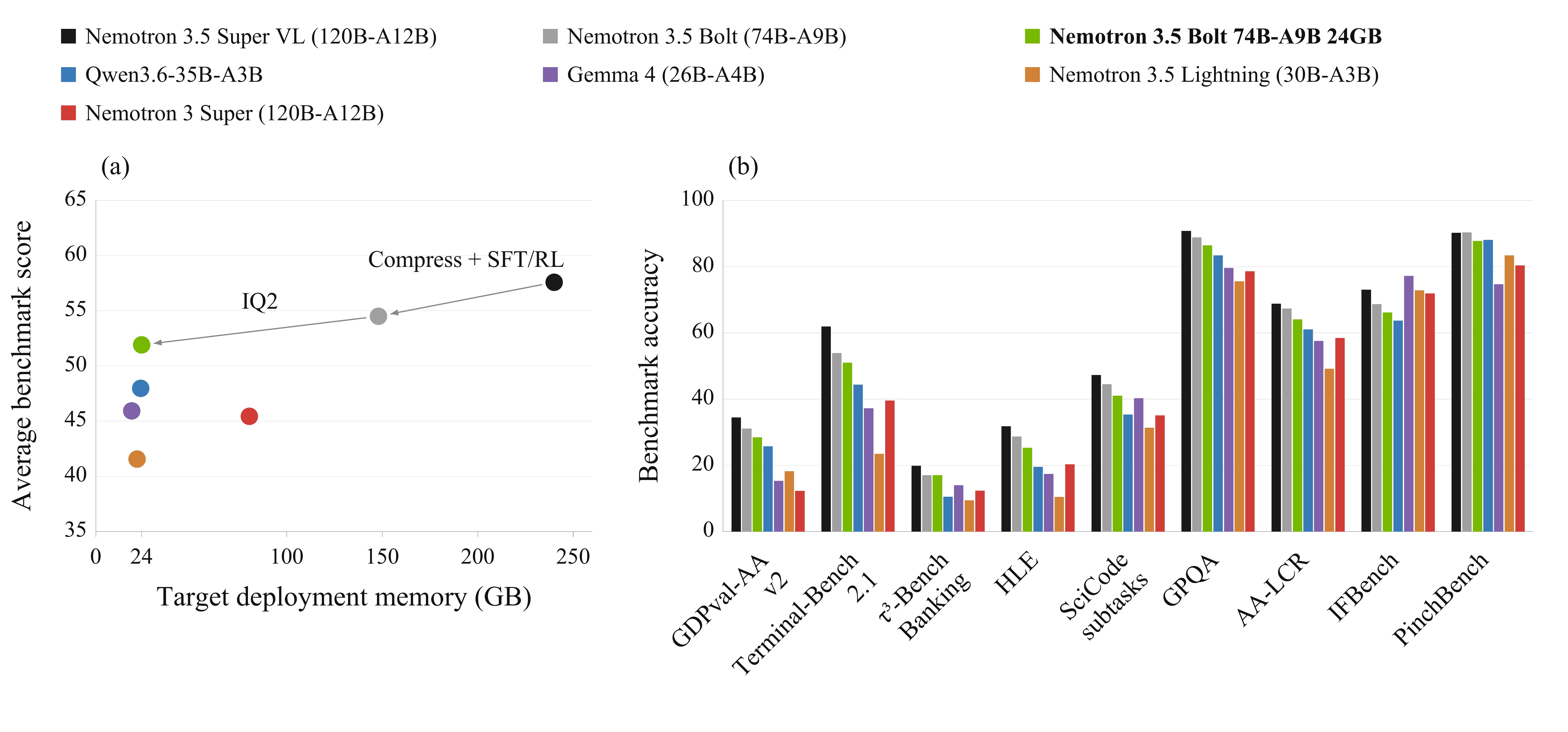}
  \caption{Per-benchmark performance of \ourmodel compared with the Nemotron 3.5 Super VL parent and community models. (a) Mean score across benchmarks versus deployment memory. (b) Individual benchmark scores. Accuracy scores consider BF16 precision for all the models, except the \ourmodel which is 2-bit. Deployment size considers all models at NVFP4, except Nemotron 3.5 VL (BF16), Nemotron 3.5 Bolt (BF16 - SFT + RL healing on compressed checkpoint) and \ourmodel (2 bit).}
  \label{fig:qad_final}
\end{figure}

Figure~\ref{fig:qad_final} compares \ourmodel with the Nemotron 3.5 Super VL parent checkpoint (120B-A12B) and other reference models.

\section{Speculative Decoding}
\label{sec:speculation}

% LiLiCorr chapter for the Nemotron 3.5 Super technical report. DRAFT.
% Bib keys are placeholders; align with the report's .bib (Maor).
% \pending{...} marks content still to be decided or measured; remove the macro before merging.
\providecommand{\pending}[1]{\textcolor{red}{\textbf{[PENDING:} #1\textbf{]}}}
\providecommand{\tbd}{\textcolor{red}{\textbf{TBD}}}

We provide a LiLiCorr ~\citep{rusanovsky2026lilicorrlightweightlikelihoodcorrelation} parallel drafter for \ourmodel{} to accelerate local inference at low concurrency.
In this regime, weight reads limit performance, and verifying a block of draft tokens costs nearly the same as generating a single token. Parallel drafting proposes several tokens in one forward pass, amortizing the weight-loading cost over several generated tokens, and significantly speeding up decoding speed. 
% LiLiCorr improves compatibility between draft tokens with a lightweight correlation head evaluated once per block.

LiLiCorr accelerates speculative decoding by jointly scoring token candidates across an entire draft in one lightweight network pass, improving on DFlash’s ~\citep{dflash2025} independent token selection while avoiding DSpark’s token-by-token neural correction, producing more coherent drafts with little overhead.

In our experiments, we have observed LiLiCorr to significantly outperform DSpark ~\citep{cheng2026dspark} both in short-context evaluation where it gets $10\%$ better AR at the same amount of training and in long-context performance where LiLiCorr gets nearly $20\%$ better throughput.

\subsection{LiLiCorr training}
\label{sec:lilicorr_training}

\paragraph{Introduction.}
DFlash~\citep{dflash2025} uses a diffusion-style Transformer to predict a block of future tokens in parallel. Although its attention connects the positions, it selects each token independently from the predicted distributions. These tokens can be individually plausible but inconsistent as a sequence.

DSpark~\citep{cheng2026dspark} addresses this problem with a lightweight neural correction applied one token at a time. This, however, breaks the promise of fully parallel drafting. 

To improve on this, LiLiCorr replaces a sequential DSpark head with a fully parallel correlation head and a compute-free sequential selector. Our configuration uses a block of $B$ positions: the last verified token and $B - 1$ future positions. At each future position, LiLiCorr retains the $K$ most likely tokens, giving $(B - 1) * K$ candidates in total.

A two-layer Transformer then processes all candidates together, incorporating their token embeddings, drafter hidden states, probability features, and position information. Its outputs are combined with target context to produce two vectors per candidate. 

The cosine similarity between these vectors measures compatibility between candidates at adjacent positions. Each score therefore reflects information from the entire candidate set.

All of the aforementioned operations are fully parallel. After they are completed, we select draft tokens from left to right using the precomputed compatibility scores. This selection requires no additional computation.

See~\citet{rusanovsky2026lilicorrlightweightlikelihoodcorrelation} for the full method.

\paragraph{Setup.} We train the drafter body and the LiLiCorr head jointly in the NVIDIA Model Optimizer (ModelOpt)~\citep{nvidia-modelopt}, in a streaming configuration suited to a target of this size. Target replicas served through vLLM~\citep{kwon2023efficient} stream the hidden states of five captured target layers, together with the final hidden state, to the trainers. The target uses the IQ2 weights folded into BF16. 

The drafter uses a block of $B=8$ positions, i.e., $7$ drafted slots. Each of the five layers of the drafter body wraps the attention and MLP sublayers in a grouped dynamic convolution~\citep{dflash2_2026}, which lets neighboring slots exchange local information while the block remains parallel. The resulting drafter has $1.13$B parameters, excluding shared target embeddings and the LM head. The two-layer LiLiCorr head accounts for $37$M of these parameters.

\paragraph{Data.} The training data draw on the Nemotron post-training sources listed in Table~\ref{tab:lilicorr_data}. For each valid assistant target turn, we generate a response conditioned on the preceding conversation, with reasoning enabled, temperature $1.0$, and top-$p$ $0.95$. The supervision mask selects only its generated response. We preprocess the corpus for a $4{,}096$-token training window. After filtering, we retain $7.55$M samples, totaling $19.14$B input tokens, of which $14.06$B are selected by the supervision mask.

\paragraph{Hyperparameters.} We train and serve the model with sliding-window attention over the most recent $1{,}024$ positions in every drafter layer, so that query-key distances never exceed those seen in training. Additionally, we use attention sinks since we've observed an improved acceptance rate when they are enabled. Further hyperparameters are presented in Table~\ref{tab:lilicorr_hparams}.

\begin{table*}[t]
  \centering
  \footnotesize
  \setlength{\tabcolsep}{3pt}

  \begin{minipage}[t]{0.46\textwidth}
    \vspace{0pt}
    \centering
    \begin{tabularx}{\linewidth}{@{}>{\raggedright\arraybackslash}Xr@{}}
      \toprule
      Dataset & Prompts \\
      \midrule
      \href{https://huggingface.co/datasets/nvidia/Nemotron-Competitive-Programming-v1}{Competitive-Programming-v1} & $1{,}174{,}694$ \\
      \href{https://huggingface.co/datasets/nvidia/Nemotron-SFT-Agentic-v2}{SFT-Agentic-v2} & $991{,}900$ \\
      \href{https://huggingface.co/datasets/nvidia/Nemotron-SFT-Instruction-Following-Chat-v2}{SFT-Instruction-Following-Chat-v2} & $929{,}237$ \\
      \href{https://huggingface.co/datasets/nvidia/Nemotron-Post-Training-Dataset-v2}{Post-Training-Dataset-v2} & $800{,}000$ \\
      \href{https://huggingface.co/datasets/nvidia/Nemotron-RL-Super-Training-Blends}{RL-Super-Training-Blends} & $479{,}303$ \\
      \href{https://huggingface.co/datasets/nvidia/Nemotron-SFT-OpenCode-v1}{SFT-OpenCode-v1} & $460{,}254$ \\
      \href{https://huggingface.co/datasets/nvidia/Nemotron-Instruction-Following-Chat-v1}{Instruction-Following-Chat-v1} & $430{,}978$ \\
      \href{https://huggingface.co/datasets/nvidia/Nemotron-SFT-SWE-v2}{SFT-SWE-v2} & $256{,}254$ \\
      \href{https://huggingface.co/datasets/nvidia/Nemotron-Science-v1}{Science-v1} & $226{,}334$ \\
      \href{https://huggingface.co/datasets/nvidia/Nemotron-SFT-Math-v3}{SFT-Math-v3} & $200{,}000$ \\
      \href{https://huggingface.co/datasets/nvidia/Nemotron-SFT-Multilingual-v1}{SFT-Multilingual-v1} & $200{,}000$ \\
      \href{https://huggingface.co/datasets/nvidia/Nemotron-SFT-Competitive-Programming-v2}{SFT-Competitive-Programming-v2} & $175{,}808$ \\
      \midrule
      Total & $6{,}324{,}762$ \\
      \bottomrule
    \end{tabularx}
    \caption{Training prompt sources. All dataset names have the prefix \texttt{Nemotron-}. Split details are provided in the appendix.}
    \label{tab:lilicorr_data}
  \end{minipage}
  \hfill
  \begin{minipage}[t]{0.52\textwidth}
    \vspace{0pt}
    \centering
    \begin{tabularx}{\linewidth}{@{}>{\raggedright\arraybackslash}X>{\raggedright\arraybackslash}X@{}}
      \toprule
      Setting & Value \\
      \midrule
      Attention window; body KV heads & $1024$; $2$ \\
      Block size; drafted slots; candidates & $8$; $7$; $8$ \\
      Head layers, width, heads, FFN & $2$, $1024$, $8$, $2048$ \\
      Body convolution (kernel, group) & $2$, $16$ \\
      Cosine scale $\sigma$; position decay $\beta$ & $8$; $4$ \\
      Body loss; head loss & Distillation; $0.05\mathcal{L}_{\mathrm{CE}}+0.45\mathcal{L}_{\mathrm{cal}}$ \\
      Optimizer; learning rate & AdamW; $8\times10^{-4}$ \\
      Schedule (warmup, decay steps) & Warmup-Stable-Decay ($1000$, $10{,}000$) \\
      Weight decay; gradient clipping & $0$; $1.0$ \\
      Batch; sequence length; blocks/sequence; epochs & $64$; $4096$; $512$; $2$ \\
      Precision & BF16, FP32 master weights \\
      Hardware & $96\times$ GB200 ($64$ train, $32$ target) \\
      \bottomrule
    \end{tabularx}
    \caption{Training configuration of LiLiCorr for \ourmodel{}.}
    \label{tab:lilicorr_hparams}
  \end{minipage}
\end{table*}

\subsection{Drafter Quantization}
\label{sec:quant:drafter}

Profiling shows that linear projection layers of the drafter contribute the most to both latency and size of the model. Therefore, we picked those as our main target for quantization.

We apply weight-only PTQ without calibration data. Quantizing linear layers present in the DFlash backbone to NVFP4 while keeping activations at BF16 precision results in an $11\%$ throughput improvement on DGX Spark.  

In the LiLiCorr head, we quantize the context, hidden-state, and token projections (\texttt{context\_proj}, \texttt{pass\_hidden\_proj}, \texttt{token\_proj}), attention output projection (\texttt{attn\_out\_proj}), and MLP projections (\texttt{mlp}). We have empirically found these to allow for the least acceptance rate drop.

\subsection{Drafter Evaluation}
\label{sec:lilicorr_eval}

We evaluate on the qualitative split of SPEED-Bench~\citep{2026speedbench}, which comprises $880$ prompts, $80$ in each of $11$ categories. The drafter is served with the LiLiCorr integration of vLLM~\citep{kwon2023efficient}\footnote{\url{https://github.com/vllm-project/vllm/pull/57934}} with a draft length of $7$ and a concurrency of $64$. We report the per-sample acceptance length (AL), including the token contributed by the target, averaged over each request and then over requests. $temp=0$ uses greedy decoding. $temp=1$, with top-$p=0.95$, uses probabilistic draft sampling. For both settings we report the mean and standard deviation over five seeds.

Since quantization is expected to degrade final AL, we report both BF16 and NVFP4 drafter performance in table~\ref{tab:lilicorr_eval}. Results are reported per category. We can clearly see that quantization does not degrade performance in any meaningful way while reducing latency by a reasonable amount.

In addition, to compare performance in real-world agentic tasks, we estimate natural AL on AA-AgentPerf Local\footnote{\url{https://github.com/ArtificialAnalysis/aa-agentperf-local}} using DSpark and LiLiCorr which share the same DFlash backbone architecture but have different heads and are trained over the same number of steps. Results in Table~\ref{tab:bolt-estimated-al} clearly show that LiLiCorr outperforms DSpark on real-world scenarios.

\begin{table}[t]
    \centering
    \small
    \begin{tabular}{lcccc}
      \toprule
      & \multicolumn{2}{c}{BF16 drafter}
      & \multicolumn{2}{c}{NVFP4 drafter} \\
      \cmidrule(lr){2-3} \cmidrule(lr){4-5}
      Category & $temp=0$ & $temp=1$ & $temp=0$ & $temp=1$ \\
      \midrule
      coding        & $5.738 \pm 0.027$ & $5.275 \pm 0.085$ & $5.693 \pm 0.020$ & $5.180 \pm 0.100$ \\
      rag           & $5.469 \pm 0.044$ & $5.117 \pm 0.018$ & $5.457 \pm 0.015$ & $5.083 \pm 0.042$ \\
      multilingual  & $5.565 \pm 0.069$ & $5.029 \pm 0.040$ & $5.529 \pm 0.018$ & $4.955 \pm 0.135$ \\
      summarization & $5.406 \pm 0.353$ & $4.863 \pm 0.047$ & $5.215 \pm 0.050$ & $4.786 \pm 0.043$ \\
      math          & $4.939 \pm 0.038$ & $4.516 \pm 0.097$ & $4.947 \pm 0.013$ & $4.457 \pm 0.094$ \\
      reasoning     & $4.796 \pm 0.016$ & $4.447 \pm 0.070$ & $4.799 \pm 0.023$ & $4.450 \pm 0.050$ \\
      qa            & $4.326 \pm 0.049$ & $3.900 \pm 0.037$ & $4.335 \pm 0.040$ & $3.923 \pm 0.058$ \\
      stem          & $4.119 \pm 0.049$ & $3.645 \pm 0.057$ & $4.092 \pm 0.012$ & $3.629 \pm 0.070$ \\
      humanities    & $4.073 \pm 0.043$ & $3.522 \pm 0.079$ & $3.986 \pm 0.035$ & $3.492 \pm 0.083$ \\
      writing       & $3.725 \pm 0.027$ & $3.421 \pm 0.052$ & $3.698 \pm 0.041$ & $3.358 \pm 0.050$ \\
      roleplay      & $3.943 \pm 0.414$ & $3.428 \pm 0.027$ & $3.747 \pm 0.035$ & $3.418 \pm 0.057$ \\
      \midrule
      Overall       & $4.736 \pm 0.078$ & $4.288 \pm 0.036$ & $4.681 \pm 0.010$ & $4.248 \pm 0.050$ \\
      \bottomrule
    \end{tabular}
    \caption{Mean request acceptance length (AL) of LiLiCorr
    on the SPEED-Bench qualitative split, with draft length $7$
    and concurrency $64$ in vLLM.
    The step-$208{,}000$ drafter uses attention sinks and
    $1{,}024$-token sliding-window attention.
    The verifier is iteration $6226$.
    Each category contains $80$ requests, with $880$ requests overall.
    Results use five seeds.
    $temp=0$ uses greedy decoding.
    $temp=1$ uses probabilistic draft sampling with top-$p=0.95$.}
    \label{tab:lilicorr_eval}
\end{table}

\begin{table}[ht]
\centering
\caption{Estimated acceptance length (AL) on AA AgentPerf Local with natural acceptance.}
\label{tab:bolt-estimated-al}
\begin{tabular}{lc}
\hline
\textbf{Model type} & \textbf{Estimated AL} \\
\hline
LiLiCorr & 4.18 \\
DSpark & 3.68 \\
\hline
\end{tabular}

\smallskip
\end{table}

\section{Inference Serving}
\label{sec:serving}
We support \ourmodel{} across two inference paths: llama.cpp with GGUF
checkpoints, and vLLM with ModelOpt-exported checkpoints. These paths address
different deployment needs: llama.cpp adds model and quantization support
for local and edge systems, while vLLM supports high-throughput serving
across accelerator platforms.

To get the best possible performance, we had to tackle different challenges in each framework. llama.cpp lacked support for LiLiCorr and contained inefficiencies in the SSM scan operation\footnote{\url{https://github.com/ggml-org/llama.cpp/pull/29807}; \url{https://github.com/ggml-org/llama.cpp/pull/28717}; \url{https://github.com/ggml-org/llama.cpp/pull/30506}} while vLLM previously did not support efficient execution of the GGUF IQ2\_XXS and Q8\_0 formats, especially on consumer-grade hardware, most commonly associated with the SM120 and SM121 architectures found in systems such as DGX Spark, RTX Spark, and RTX 5090.

In our profiling, we identified that while base-model execution was on-par between llama.cpp and vLLM when using the vllm-gguf-plugin\footnote{\url{https://github.com/vllm-project/vllm-gguf-plugin}}, speculative decoding was $70\%$ worse than in llama.cpp. To improve vLLM performance, we've added kernels optimized specifically for SM12x platforms for the numeric formats used throughout this work (IQ2\_XXS and Q8\_0) to b12x, which is natively integrated into vLLM\footnote{\url{https://github.com/local-inference-lab/b12x};\url{https://github.com/vllm-project/vllm/pull/59604}}. This gave us a significant boost in speculative decoding leading to improved performance even over llama.cpp. 

\section{Evaluation and performance}
\label{sec:evaluation}

\begin{table}[htbp]
\centering
\caption{Accuracy benchmark comparison across Nemotron, Qwen, and Gemma models}
\label{tab:bolt-bf16-iq2-qwen-comparison}
\resizebox{\textwidth}{!}{%
\begin{tabular}{lccccccc}
\toprule
\textbf{Benchmark} &
\shortstack{\textbf{Nemotron 3.5 Super}\\\textbf{120B-A12B-BF16}} &
\shortstack{\textbf{Nemotron 3.5 Bolt}\\\textbf{74B-A9B-BF16}} &
\shortstack{\textbf{Nemotron 3.5 Bolt}\\\textbf{74B-A9B-24GB}} &
\shortstack{\textbf{Qwen 3.6}\\\textbf{35B-A3B-BF16}} &
\shortstack{\textbf{Gemma4}\\\textbf{26B-A4B}} &
\shortstack{\textbf{Nemotron 3.5 Lightning}\\\textbf{30B-A3B}} &
\shortstack{\textbf{Nemotron 3 Super}\\\textbf{120B-A12B-BF16}} \\
\midrule
GDPVal-AA-V2               & 34.48 & 31.13 & 28.47 & 25.75 & 15.33 & 18.26 & 12.27 \\
Terminal-Bench 2.1         & 61.94 & 53.93 & 50.98 & 44.38 & 37.22 & 23.46 & 39.61 \\
Tau3 Banking               & 19.85 & 17.01 & 17.01 & 10.52 & 14.02 & 9.48  & 12.37 \\
HLE (no tools)             & 31.84 & 28.73 & 25.30 & 19.56 & 17.42 & 10.47 & 20.30 \\
AA-Omniscience accuracy    & 25.98 & 21.80 & 16.90 & 19.47 & 22.17 & 16.63 & 26.68 \\
SciCode                    & 47.27 & 44.49 & 41.02 & 35.33 & 40.28 & 31.38 & 35.11 \\
GPQA Diamond               & 90.78 & 88.83 & 86.43 & 83.40 & 79.61 & 75.57 & 78.60 \\
AA-LCR                     & 68.81 & 67.38 & 64.06 & 61.06 & 57.56 & 49.19 & 58.44 \\
AA-Omniscience non-hallucination & 69.24 & 56.14 & 61.27 & 65.85 & 14.22 & 68.81 & 14.64 \\
\addlinespace
PinchBench                 & 90.25 & 90.35 & 87.82 & 88.07 & 74.70 & 83.43 & 80.36 \\
MCP-Atlas                  & 51.20 & 55.60 & 52.20 & 51.2\textsuperscript{*} & --    & 46.4    & 52.8 \\
BrowseComp                 & 71.50 & 70.00 & 66.75 & 48.74 & 26.30 & 36.81 & 22.77 \\
WebArena (vision)          & 57.40 & 50.25 & 50.86 & 36.60 & 38.2    & --    & -- \\
\addlinespace
IFBench (instruction loose)            & 73.04 & 68.71 & 66.17 & 63.71 & 77.25 & 72.88 & 71.92 \\
\bottomrule
\end{tabular}
}
\parbox{\textwidth}{\footnotesize
\textsuperscript{*}The reported MCP-Atlas score for Qwen3.6-35B-A3B-BF16 is 62.8; however, we were unable to reproduce that result in our evaluation, which yielded 51.2.
}
\end{table}

For MCP-Atlas evaluations, we used the official eval \mbox{repo}\footnote{\url{https://github.com/scaleapi/mcp-atlas}} with reasoning-on, and default inference parameters on the public 500 tasks. 

For WebArena evaluations, we used an internal parallelized harness
on the public 812-task WebArena benchmark\footnote{\url{https://github.com/web-arena-x/webarena}},
with reasoning enabled, screenshot-based browser interaction,
and vLLM inference using our default inference settings
(temperature 1.0 and top-$p$ 0.95).
We allowed up to 100 agent steps per task. Tasks were scored
using an internal implementation of WebArena's string, URL,
and programmatic webpage checks.
For BrowseComp, we used the Exa web search API for this evaluation. 
For all other benchmarks, we used the same evaluation setup as the Super 3.5 VL parent model.

\subsection{Performance}
We measure the final model performance with two complementary benchmarks: AIPerf SPEED-Bench-Coding\footnote{\url{https://docs.nvidia.com/aiperf/tutorials/datasets-inputs/profile-with-speed-bench-dataset}} for throughput with controlled ISL, and AA-AgentPerf Local\footnote{\url{https://github.com/ArtificialAnalysis/aa-agentperf-local}} for end-to-end replay duration across realistic heterogeneous agent trajectories. All measurements were done with llama.cpp on Windows 11 ARM x64, with the same RTX Spark hardware (Surface Laptop Ultra, 128GB).
\begin{table}[htbp]
\centering
\caption{Performance benchmark comparison across Nemotron, Qwen, and Gemma models}
\label{tab:model-performance}
\resizebox{\textwidth}{!}{%
\begin{tabular}{llccccc}
\hline
\textbf{Benchmark}
& \textbf{Metric}
& \textbf{\shortstack{Nemotron 3.5 Bolt 74B-A9B \\ 24GB w. LiLiCorr}}
& \textbf{\shortstack{Qwen3.6 35B-A3B \\ UD-Q4\_K\_M w. MTP}}
& \textbf{\shortstack{Gemma 4 26B-A4B \\ UD-Q4\_K\_M w. MTP}}
& \textbf{\shortstack{Qwen3.8 27B \\ UD-Q4\_K\_M w. DFlash}} \\
\hline
AIPerf SPEED-Bench-Coding
& TPS
& 83.99
& 96.63
& 100.56
& 28.26 \\
AA AgentPerf Local
& Replay Duration (seconds)
& 600.80
& 453.97
& 480.68
& 1043.28 \\
\hline
\end{tabular}%
}
\end{table}

\subsection{Deployment Memory}
\label{sec:checkpoint-size}

We estimate memory for the deployment scenario with batch size 1, context length
131072, and a Mamba prefill chunk size of 8192 tokens, using the weights, caches, and activations with corresponding 
precisions stated in Table~\ref{tab:bolt-tensor-precision}. The listed
weights occupy 22.01\,GB, and cache and activations add 0.87\,GB,
giving an estimated subtotal of 22.87\,GB before speculative
decoding.

The remaining 1.13\,GB of the 24\,GB target accommodates the
speculative drafter  which uses 0.96 GB. This leaves us at 23.83GB, still with 0.17GB memory available out of 24GB. The vision encoder and
projector are listed separately and excluded from this target,
which covers text-only serving.
\section{Conclusion}
We present \ourmodel to bring the capabilities of a large hybrid vision-language model to local, memory-constrained deployment. The model uses Star Elastic to create a compact backbone, SFT and RL to restore checkpoint capabilities, 2-bit expert quantization and quantization-aware distillation to reduce memory use, and speculative decoding to improve inference. Together, these components support the development of local assistants for reasoning, coding, tool use, and visual interaction.
\section*{Contributors}

We thank the following people for their invaluable contributions to \ourmodel.

Ganesh Ajjanagadde, Chris Alexiuk, Luke Alonso, Kumar Anshuman, Maor Ashkenazi, Nave Assaf, Gorkem Batmaz, Omer Belhasin, Shai Ben Nun, Akhiad Bercovich, Pritam Biswas, Caroline Blanco, Tijmen Blankevoort, Michael Boone, Christopher Borden, Kari Briski, Bryan Catanzaro, Zijia Chen, Aishwarya Chhabra, Hung-Yueh Chiang, Annamalai Chockalingam, Jonathan Cohen, Joey Conway, Olivier Delalleau, Gerardo Delgado Cabrera, Leon Derczynski, Li Ding, Will Efstratis, Michael Elad, Ellie Evans, Wes Feely, Michael Fukuyama, Gaurav Garg, Rishabh Garg, Arvind Gopalakrishnan, Shishir Goyal, Kartik Gulia, Hao Guo, Negar Habibi, Scott Hara, Ehsan Hosseini Asl, Rohit Jena, Udi Karpas, Bilal Kartal, Mayank Kaushik, Manoj Kilaru, Mackenzie Kosseim, Oleksii Kuchaiev, Matthieu Le, Raymond Lee, Jian Leong, Jonathan Li, Sheldon Liang, Mingjie Liu, Eileen Long, Katherine Luna, Chenjie Luo, Mingyuan Ma, Arsh Maknojia, Richard Mazzarese, Jessica Medina, Puhui Meng, Yoav Miron, Pavlo Molchanov, David Mosallanezhad, Chetan Mungekar, Marina Neseem, Oluwatobi Olabiyi, Jason Paul, Bhavesh Pawar, Seth Poulos, Anav Prasad, Meredith Price, Sweta Priyadarshi, Haifeng Qian, Vinay Raman, Kaustubh Rao, Hayley Ross, Matan Rusanovsky, Sudeep Sabnis, Karan Sapra, Micah Schaffer, Schmitty, Alex Settle, Ashton Sharabiani, Haran Kumar Shiv Kumar, Rameshwar Shivbhakta, Guillermo Siman, Barnaby Simkin, Oliver Simons, Andrii Skliar, Padmavathy Subramanian, Yoshi Suhara, Ali Taghibakhshi, Nima Tajbakhsh, Andrew Tao, Roy Uziel, Prasoon Varshney, Jose Vergara, Huy Vu, Mengru Wang, Rochelle Woods, Pinky Xu, Dina Yared, Chenhan Yu, Chaosen Zhang, Jian Zhang, Yiheng Zhang, Ran Zilberstein.

We also thank the rest of the team that made Super 3.5 VL a success.

\newpage
\FloatBarrier
\bibliographystyle{references}
\bibliography{references}

@article{jang2016categorical,
  title={Categorical reparameterization with gumbel-softmax},
  author={Jang, Eric and Gu, Shixiang and Poole, Ben},
  journal={arXiv preprint arXiv:1611.01144},
  year={2016}
}

@inproceedings{lasby2026reap,
  title={Reap the experts: Why pruning prevails for one-shot moe compression},
  author={Lasby, Mike and Lazarevich, Ivan and Sinnadurai, Nish and Lie, Sean and Ioannou, Yani and Thangarasa, Vithursan},
  booktitle={International Conference on Learning Representations},
  volume={2026},
  pages={146883--146912},
  year={2026}
}

@misc{taghibakhshi2026starelastic_arxiv,
  title = {{Star Elastic}: Many-in-One Reasoning {LLMs} with Efficient Budget Control},
  author = {Ali Taghibakhshi and Ruisi Cai and Saurav Muralidharan and Sharath Turuvekere Sreenivas and Aditya Vavre and Ameya Sunil Mahabaleshwarkar and Bilal Kartal and Sheldon Liang and Marcin Chochowski and Zijia Chen and Akhiad Bercovich and Ran Zilberstein and Ran El-Yaniv and Yonatan Geifman and Daniel Korzekwa and Yoshi Suhara and Oluwatobi Olabiyi and Ashwath Aithal and Nima Tajbakhsh and Pavlo Molchanov},
  year = {2026},
  eprint = {2605.07182},
  archivePrefix = {arXiv},
  primaryClass = {cs.LG},
  doi = {10.48550/arXiv.2605.07182},
  url = {https://arxiv.org/abs/2605.07182v1}
}

@misc{xin2026quantizationawaredistillationnvfp4inference,
      title={Quantization-Aware Distillation for NVFP4 Inference Accuracy Recovery}, 
      author={Meng Xin and Sweta Priyadarshi and Jingyu Xin and Bilal Kartal and Aditya Vavre and Asma Kuriparambil Thekkumpate and Zijia Chen and Ameya Sunil Mahabaleshwarkar and Ido Shahaf and Akhiad Bercovich and Kinjal Patel and Suguna Varshini Velury and Chenjie Luo and Zhiyu Cheng and Jenny Chen and Chen-Han Yu and Wei Ping and Oleg Rybakov and Nima Tajbakhsh and Oluwatobi Olabiyi and Dusan Stosic and Di Wu and Song Han and Eric Chung and Sharath Turuvekere Sreenivas and Bryan Catanzaro and Yoshi Suhara and Tijmen Blankevoort and Huizi Mao},
      year={2026},
      eprint={2601.20088},}

@misc{llamacpp_tensor_encoding,
  author       = {{ggml-org}},
  title        = {llama.cpp: Tensor Encoding Schemes},
  howpublished = {\url{https://github.com/ggml-org/llama.cpp/wiki/Tensor-Encoding-Schemes}}
}

@misc{liu2025paretoqimprovingscalinglaws,
      title={ParetoQ: Improving Scaling Laws in Extremely Low-bit LLM Quantization}, 
      author={Zechun Liu and Changsheng Zhao and Hanxian Huang and Sijia Chen and Jing Zhang and Jiawei Zhao and Scott Roy and Lisa Jin and Yunyang Xiong and Yangyang Shi and Lin Xiao and Yuandong Tian and Bilge Soran and Raghuraman Krishnamoorthi and Tijmen Blankevoort and Vikas Chandra},
      year={2025},
      eprint={2502.02631},
      archivePrefix={arXiv},
      primaryClass={cs.LG},
      url={https://arxiv.org/abs/2502.02631}, 
}

@misc{panferov2025queststabletrainingllms,
      title={QuEST: Stable Training of LLMs with 1-Bit Weights and Activations}, 
      author={Andrei Panferov and Jiale Chen and Soroush Tabesh and Roberto L. Castro and Mahdi Nikdan and Dan Alistarh},
      year={2025},
      eprint={2502.05003},
      archivePrefix={arXiv},
      primaryClass={cs.LG},
      url={https://arxiv.org/abs/2502.05003}, 
}

@misc{hinton2015distillingknowledgeneuralnetwork,
      title={Distilling the Knowledge in a Neural Network}, 
      author={Geoffrey Hinton and Oriol Vinyals and Jeff Dean},
      year={2015},
      eprint={1503.02531},
      archivePrefix={arXiv},
      primaryClass={stat.ML},
      url={https://arxiv.org/abs/1503.02531}, 
}

@misc{gu2026minillmonpolicydistillationlarge,
      title={MiniLLM: On-Policy Distillation of Large Language Models}, 
      author={Yuxian Gu and Li Dong and Furu Wei and Minlie Huang},
      year={2026},
      eprint={2306.08543},
      archivePrefix={arXiv},
      primaryClass={cs.CL},
      url={https://arxiv.org/abs/2306.08543}, 
}

@inproceedings{wen-etal-2023-f,
    title = "f-Divergence Minimization for Sequence-Level Knowledge Distillation",
    author = "Wen, Yuqiao  and
      Li, Zichao  and
      Du, Wenyu  and
      Mou, Lili",
    editor = "Rogers, Anna  and
      Boyd-Graber, Jordan  and
      Okazaki, Naoaki",
    booktitle = "Proceedings of the 61st Annual Meeting of the {A}ssociation for {C}omputational {L}inguistics (Volume 1: Long Papers)",
    month = jul,
    year = "2023",
    address = "Toronto, Canada",
    publisher = "Association for Computational Linguistics",
    url = "https://aclanthology.org/2023.acl-long.605/",
    doi = "10.18653/v1/2023.acl-long.605",
    pages = "10817--10834"
}

@misc{sanh2020distilbertdistilledversionbert,
      title={DistilBERT, a distilled version of BERT: smaller, faster, cheaper and lighter}, 
      author={Victor Sanh and Lysandre Debut and Julien Chaumond and Thomas Wolf},
      year={2020},
      eprint={1910.01108},
      archivePrefix={arXiv},
      primaryClass={cs.CL},
      url={https://arxiv.org/abs/1910.01108}, 
}

@misc{rusanovsky2026lilicorrlightweightlikelihoodcorrelation,
  title={LiLiCorr: Lightweight Likelihood Correlation of Parallel Drafts for Speculative Decoding},
  author={Matan Rusanovsky and Yoav Miron and Roy Uziel and Omer Belhasin and Hao Guo and Ran Zilberstein and Maor Ashkenazi and Michael Elad},
  year={2026},
  eprint={2608.20530},
  archivePrefix={arXiv},
  primaryClass={cs.CL},
  url={https://arxiv.org/abs/2608.20530}
}

@inproceedings{dflash2025,
  title={{DF}lash: Block Diffusion for Flash Speculative Decoding},
  author={Jian Chen and Yesheng Liang and Zhijian Liu},
  booktitle={Forty-third International Conference on Machine Learning},
  year={2026},
  url={https://openreview.net/forum?id=Oz335dV48X}
}

@article{cheng2026dspark,
  title={DSpark: Confidence-Scheduled Speculative Decoding with Semi-Autoregressive Generation},
  author={Cheng, Xin and Yu, Xingkai and Shao, Chenze and Li, Jiashi and Xiong, Yunfan and Qian, Yi and Zhu, Jiaqi and Ma, Shirong and Zhang, Xiaokang and Ye, Jiasheng and others},
  journal={arXiv preprint arXiv:2607.05147},
  year={2026}
}

@misc{nvidia-modelopt,
  author={{NVIDIA Corporation}},
  title={{NVIDIA Model Optimizer}},
  howpublished={\url{https://github.com/NVIDIA/Model-Optimizer}},
  year={2024--2026},
  note={GitHub repository}
}

@inproceedings{kwon2023efficient,
  title={Efficient Memory Management for Large Language Model Serving with PagedAttention},
  author={Woosuk Kwon and Zhuohan Li and Siyuan Zhuang and Ying Sheng and Lianmin Zheng and Cody Hao Yu and Joseph E. Gonzalez and Hao Zhang and Ion Stoica},
  booktitle={Proceedings of the ACM SIGOPS 29th Symposium on Operating Systems Principles},
  year={2023}
}

@misc{dflash2_2026,
  title={{DFlash 2: Keep Drafting Parallel}},
  author={{Inco AI}},
  year={2026},
  month={August},
  url={https://inco.ai/blog/dflash2/}
}

@inproceedings{2026speedbench,
  title={{SPEED}-Bench: A Unified and Diverse Benchmark for Speculative Decoding},
  author={Talor Abramovich and Maor Ashkenazi and Izzy Putterman and Benjamin Chislett and Tiyasa Mitra and Bita Darvish Rouhani and Ran Zilberstein and Yonatan Geifman},
  booktitle={Forty-third International Conference on Machine Learning},
  year={2026},
  url={https://openreview.net/forum?id=Rl2uQlCoQX}
}
\end{document}